\documentclass[letterpaper, 10 pt, conference]{ieeeconf}
\IEEEoverridecommandlockouts
\usepackage{graphics}
\usepackage{subcaption}
\usepackage{times}
\usepackage{amsmath}
\usepackage{amssymb}
\usepackage{amsfonts}
\usepackage{mathrsfs}
\usepackage{mathtools}
\usepackage{booktabs}
\usepackage{diagbox}
\usepackage{makecell}
\DeclarePairedDelimiter{\ceil}{\lceil}{\rceil}

\title{\LARGE \bf
Spatio-Temporal Attention Network for Persistent Monitoring of Multiple Mobile Targets}

\author{Yizhuo Wang$^{1}$, Yutong Wang$^{1}$, Yuhong Cao$^{1}$, Guillaume Sartoretti$^{1}$$^{\dagger}$%
\thanks{$\dagger$ Corresponding author, to whom correspondence should be addressed.}%
\thanks{$^{1}$Authors are with the Department of Mechanical Engineering, College of Design and Engineering, National University of Singapore 
{\tt\small \{wy98,e0576114,caoyuhong\}@u.nus.edu, mpegas@nus.edu.sg}%
}}

\begin{document}
\maketitle
\thispagestyle{empty}
\pagestyle{empty}


\begin{abstract}

This work focuses on the persistent monitoring problem, where a set of targets moving based on an unknown model must be monitored by an autonomous mobile robot with a limited sensing range.
To keep each target's position estimate as accurate as possible, the robot needs to adaptively plan its path to (re-)visit all the targets and update its belief from measurements collected along the way.
In doing so, the main challenge is to strike a balance between \textit{exploitation}, i.e., re-visiting previously-located targets, and \textit{exploration}, i.e., finding new targets or re-acquiring lost ones.
Encouraged by recent advances in deep reinforcement learning, we introduce an attention-based neural solution to the persistent monitoring problem, where the agent 
can learn the inter-dependencies between targets, i.e., their spatial and temporal correlations, conditioned on past measurements.
This endows the agent with the ability to determine which target, time, and location to attend to across multiple scales, which we show also helps relax the usual limitations of a finite target set.
We experimentally demonstrate that our method outperforms other baselines in terms of number of targets visits and average estimation error in complex environments.
Finally, we implement and validate our model in a drone-based simulation experiment to monitor mobile ground targets in a high-fidelity simulator.

\end{abstract}


\section{Introduction}

Persistent monitoring refers to problems where autonomous robots equipped with onboard sensors are tasked with collecting data to maintain an accurate and up-to-date awareness of a given environment.
Compared to stationary sensor systems, mobile robots can establish a dynamic sensor network, providing greater flexibility in deployment and handling a broader spectrum of tasks, particularly in scenarios such as wildlife habitat monitoring~\cite{tokekar2013tracking}, traffic surveillance~\cite{booth2020target}, and search-and-rescue missions~\cite{kashino2020hybrid, sung2017algorithm}.
In such cases, it may not be possible to infer the exact dynamics of the environment due to the stochastic nature of the targets' movements, limiting the use of a fixed sensor network.

In this paper, we specifically investigate the problem of controlling a mobile robot to persistently monitor a set of mobile targets moving over a given domain.
Our agent is equipped with a binary sensor with limited Field-of-View (FoV) \cite{coffin2022multiagent} (e.g., a camera with a visual classifier for target detection), and may or may not know the number and initial positions of the targets {\it a priori}.
The underlying motion model/dynamics of the targets is assumed unknown.
Based on measurements along its path, the agent must build and update a time-dependent {\it belief} of each individual target location, to reason about their possible locations to frequently re-visite/-locate each of them.

\begin{figure}[t]
    \centering
    \subcaptionbox{}{\vspace{-0.15cm}
    \includegraphics[width=0.14\textwidth]{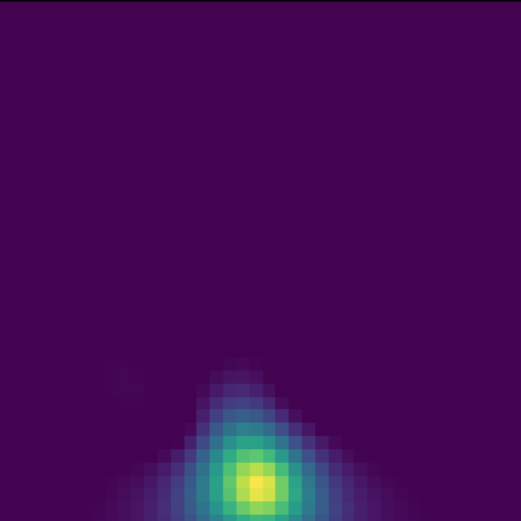}}%
    \hfill
    \subcaptionbox{}{\vspace{-0.15cm}
    \includegraphics[width=0.14\textwidth]{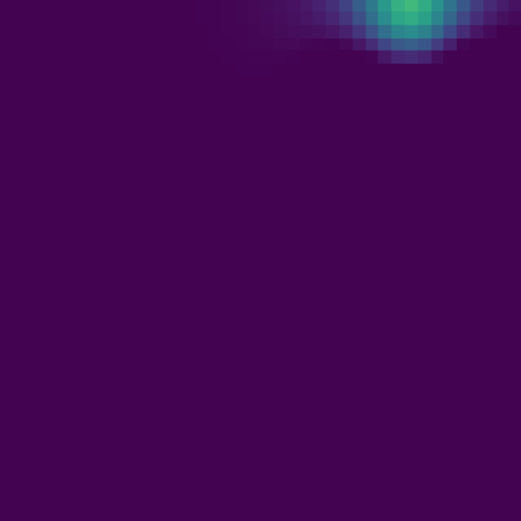}}%
    \hfill
    \subcaptionbox{}{\vspace{-0.15cm}
    \includegraphics[width=0.14\textwidth]{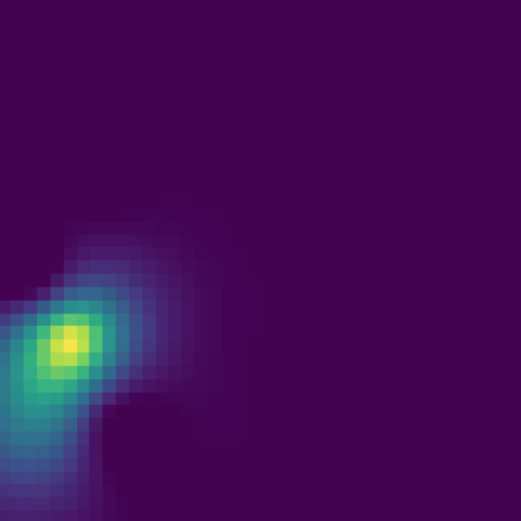}}%
    \hfill
    \subcaptionbox{}{\vspace{-0.15cm}
    \includegraphics[width=0.14\textwidth]{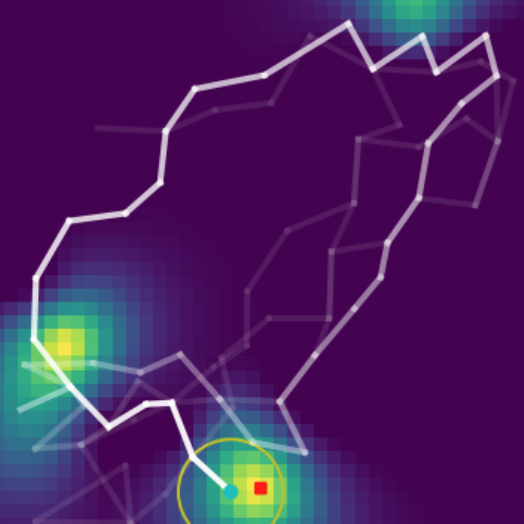}}%
    \hfill
    \subcaptionbox{}{\vspace{-0.15cm}
    \includegraphics[width=0.14\textwidth]{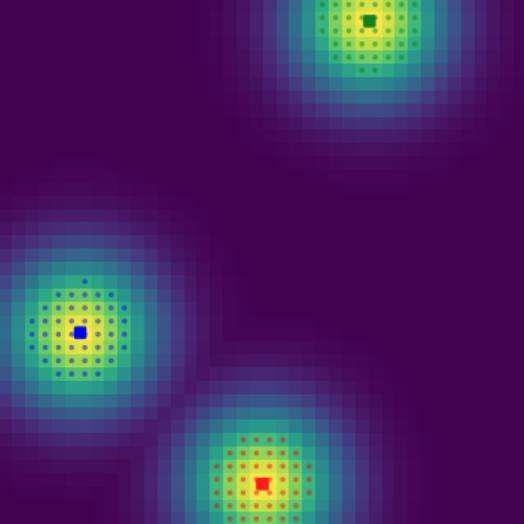}}%
    \hfill
    \subcaptionbox{}{\vspace{-0.15cm}
    \includegraphics[width=0.14\textwidth]{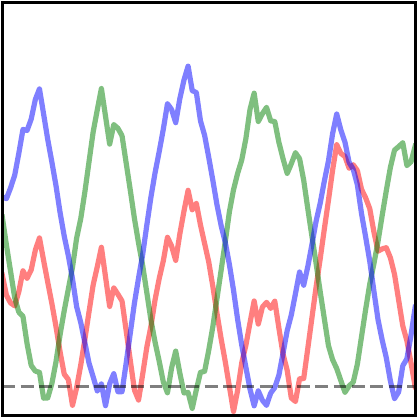}}%
    \vspace{-0.2cm}
    \caption{{\bf Persistent monitoring of three mobile targets.}
    (a-c) Current agent belief of each of the three targets.
    (d) Agent belief of all targets and executed trajectory so far (white line with growing opacity, thicker being more recent), agent position (cyan), agent FoV (yellow circle), and classified target 1 (red square) within FoV.
    (e) Target true positions and their associated target areas (colored squares and circle dots around them, both unknown to the agent).
    (f) Euclidean distance between the agent and each targets (colored lines) throughout the mission.
    Targets are only observable within the agent's FoV (here, below the dashed line).
    }
    \vspace{-0.6cm}
    \label{fig:overview}
\end{figure}

Different from monitoring an underlying vector field of interest~\cite{chen2021multiagent, nigam2008persistent}, which is typically approached as generalized coverage problem or as a variant of the orienteering problem~\cite{rossello2022informationdriven, yu2016correlated}, our agent aims to maximize {\it information gain} in the vicinity of the targets (i.e., minimizes the uncertainty over the true target positions)~\cite{popovic2020informative, cao2022catnipp}.
Since this uncertainty grows over time for targets not in direct view of the agent, the agent has to leave some of the targets temporarily untracked to relocate others and then return to those as soon as possible to avoid losing them.
If a target is lost, the agent needs to strike a balance between the cumulative penalty of losing it for good, and the potential risk of losing more targets as it attempts to relocate it, thereby making lost target recovery costly/risky.
As a result, an efficient planner must trade-off exploration and exploitation to essentially plan a route that repeatedly visits all targets' (predicted) areas in the shortest time.
A straightforward solution would be to tackle the problem as a variant of the Traveling Salesman Problem (TSP), where the agent optimizes a shortest loop to visit all targets based on its belief.
However, this approach naturally focuses solely on exploiting the belief, and thus can easily fail at relocating any lost target.
Therefore, an effective planner should associate past observations to make a context-aware decision over which target to attend to next and/or where to go to maximize the likelihood of relocating lost targets or even uncovering new ones.
However, this is very challenging for a conventional planner, due to the problem's combinatorial nature~\cite{hall2022optimal, rezazadeh2020submodular}, e.g., the unrestricted moving-target TSP has notably been shown to be NP-hard even with only two targets~\cite{helvig2003movingtarget}.
This is also why many persistent monitoring frameworks impose limitation on the number of targets~\cite{yu2017optimal, pinto2022multiagent}, thus inhibiting scalability to a varying number of targets during the mission.

To improve search performance and relax these limitations, we introduce a neural approach based on deep reinforcement learning (DRL) to solves the single-agent, multi-mobile-target persistent monitoring problem.
To reduce the complexity of the continuous 2D monitoring domain, we first construct a random sparse graph over the domain, on which the agent can iteratively build its path via selected nodes (waypoints) along existing edges~\cite{yu2016correlated, rezazadeh2020submodular}.
We then augment this roadmap with the agent's belief of all targets, and cast persistent monitoring as a sequential decision-making problem.
To fuse and extract the spatial and temporal information from all targets, our work augments the agent's belief with future predictions autoregressively, and introduces a spatio-temporal attention network that successively concentrates target features, history of recent beliefs, and graph information, to finally output a policy over which neighboring node to visit next (see Fig.~\ref{fig:overview}).
We experimentally demonstrate that, in complex environments, our method outperforms other baselines across a variety of metrics including average uncertainty, minimum observation times, and average estimation error.
We also conduct a realistic simulation in AirSim, where a drone is tasked with persistently monitoring set of random ground targets.


\section{Related Works}

Many variants of the persistent monitoring problem have been studied, and most methods can be categorised into either \textit{field monitoring} or \textit{target monitoring}, mainly depending on which quantity is associated with the belief constructed/updated by the agent (continuous field or discrete target(s)).
Krause et al. first optimized stationary sensor placements by maximizing the mutual information between location candidates to monitor field information (e.g., temperature, humidity)~\cite{krause2008nearoptimal}.
More recent works have dealt with larger outdoor environments (usually represented as a time-varying Gaussian Random Field), where autonomous robot(s) with onboard sensors need to sequentially take measurements at multiple positions and maintain a field value belief with estimators such as Gaussian Processes~\cite{ma2018multirobot,garg2018persistent}, Kalman Filters~\cite{rossello2022informationdriven,lan2013planning}, or Proper Orthogonal Decompositions~\cite{salam2019adaptive}.
Yu et al. also proposed a variant of the graph-based orienteering problem to monitor a spatio-temporal field with time-invariant spatial correlations~\cite{yu2016correlated}.

For target monitoring, most studies consider a finite set of stationary targets with evolving internal events/states~\cite{rezazadeh2020submodular,pinto2022multiagent}.
In particular, Yu et al. optimized the target dwelling time to reduce the static monitoring problem to a TSP problem where each target is periodically visited in a TSP loop~\cite{yu2017optimal}.
However, these methods often assume immobile targets without any uncertainty in their positions.
Sung et al. considered a realistic scenario where an agent with a limited-FoV, noisy sensor searches and tracks a number of mobile targets with unknown dynamics.
Their approach involves the use of a generalized Gaussian Mixture Probability Hypothesis Density (GM-PHD) filter, enabling simultaneous search and tracking while effectively modeling multiple targets~\cite{sung2017algorithm}.
However, their methodology is limited to a finite mission horizon and relies on estimating prior information.

\section{Problem Setup and Representation}

In this section, we formulate persistent monitoring in 2D environments, and define some notations of the binary sensor model used to track mobile targets. We also introduce our use of Gaussian Processes (GPs) for modeling the time-dependent belief of each target based on these measurements.

\subsection{Target Setup}

The objective of mobile targets persistent monitoring is to control an agent with limited sensing range to observe all targets as frequently as possible, to keep the position estimate uncertainty and estimate error as low as possible through time.
We consider a given set of $N$ mobile targets ($i \in \{1, \dots, N\}$) moving in a bounded 2D environment $\mathcal{E} \subset \mathbb{R}^2$.
To represent the temporal change associated with the location, we use $\mathcal{E}^* \subset \mathbb{R}^3$ to denote the time-varying monitoring domain, with respect to a reference time (e.g., the start of the mission).
Assuming targets to be point mass, let $y_{i,t} \in \mathcal{Y}$ be the position of target $i$ at time $t$, where $\mathcal{Y} \subseteq \mathcal{E}$ is the set of all possible target positions.
The targets move along a set of underlying trajectories that are unknown to the agent, and their position transition is subject to a velocity constraint.
For simplicity (but not a strict requirement), in this work, all targets abide by the same velocity constraints of moving at a constant speed which is also unknown to the agent.
That is, given a time interval, all targets travel the same distance along their underlying trajectories.
We denote the speed ratio between the targets and the agent as $r_v<1$.

\subsection{Sensor Setup}

In order to locate the targets, the agent takes a measurement at time $t$ at/around its current location $x_t \in \mathcal{E}$.
The agent is equipped with an accurate \textit{binary} sensor, which output measurements $z_{i,t} \in \{0,1\}$ indicating the presence of target $i$ in the sensor's FoV/footprint $S(x_t) \subseteq \mathcal{E}$~\cite{coffin2022multiagent}.
In our context, we assume a circular sensor footprint with radius 0.1, and consider a 2D unit square as search domain $\mathcal{E}=[0,1]^2$.
Note that the sensor can classify target index (e.g., output of a visual classifier, able to differentiate among targets).
As a result, each time the agent senses the environment, a $N$-dimensional measurement vector $\mathcal{Z}_t = \left\langle z_{i,t} \right\rangle_{i=1}^N$ is obtained, with $N$ the number of targets.
If target $i$ is within the sensor's FoV, a $1$ measurement will be obtained at that target's current position $y_{i,t}$, otherwise a $0$ measurement is obtained at the agent's current position $x_{i,t}$.
Each measurement is associated with the current time $t$, and formulate the timestamped measurement location $x_{i,t}^* \in \mathcal{E}^*$:
\vspace{-0.25cm}
\begin{equation}
\begin{cases}
    x_{i,t}^* = \{y_{i,t}, t\},\ z_{i,t} = 1 \quad & \text {if}\ y_{i,t} \in S(x_{t}) \\
    x_{i,t}^* = \{x_{i,t}, t\},\ z_{i,t} = 0 \quad & \text {if}\ y_{i,t} \notin S(x_{t}).
\end{cases}
\label{eq:measure}
\vspace{-0.15cm}
\end{equation}

\noindent Following this, the agent plans a trajectory $\psi$ and takes measurements along the it at a fixed frequency.
Since the agent's velocity is assumed constant, the number of measurements is only determined by the trajectory length $L(\psi)$.

\subsection{Gaussian Processes for the Agent's Belief}
Although sensor measurements are binary, it is advantageous to estimate a time-dependent continuous field of potential target positions as it interpolates the measurements and provides richer information about the process.
Thereby, the agent's belief is assumed to be a continuous function: $\zeta_i: \mathcal{E}^* \mapsto \mathbb{R}$.
Gaussian Processes (GP) have been widely used to model such spatial/spatio-temporal correlations in a probabilistic and non-parametric manner~\cite{popovic2020informative,ma2018multirobot,garg2018persistent}.
Unlike~\cite{sung2017algorithm} that only use GP Regressions to predict target trajectories, we adopt a separate GP for each target $i$, each characterized by the mean $\mu_i = \mathbb{E}[\zeta_i]$ and covariance $P_i = \mathbb{E}[(\zeta_i-\mu_i)(\zeta_i^\top-\mu_i^\top)]$ as $\mathcal{GP}_i(\mu_i,P_i)$.
The GP posterior for each target is built from a set of $n$ timestamped observed locations $\mathcal{X}_i^* \subset \mathcal{E}^*$ ($x_{i,t}^* \in \mathcal{X}_i^*$) and their corresponding measurements $\mathcal{Z}_i$ ($z_{i,t} \in \mathcal{Z}_i$) indicating the target's presence/absence.
For noiseless measurements, the mean and variance of GP are regressed as (with target index subscripts omitted)~\cite{williams2006gaussian}:
\vspace{-0.1cm}
\begin{equation}
\begin{split}
    \mu &= \mu(\mathcal{X}^\dagger)+K(\mathcal{X}^\dagger, \mathcal{X}^*)K(\mathcal{X}^*,\mathcal{X}^*)^{-1}(\mathcal{Z}-\mu(\mathcal{X}^*)),\\
    P &= K(\mathcal{X}^\dagger, \mathcal{X}^\dagger)-K(\mathcal{X}^\dagger, \mathcal{X}^*)K(\mathcal{X}^*,\mathcal{X}^*)^{-1}\\ &\quad\times
    K(\mathcal{X}^\dagger,\mathcal{X}^*)^\top,
\end{split}
\label{eq:gpr}
\vspace{-0.2cm}
\end{equation}
where $\mathcal{X}^\dagger \subset \mathcal{E^*}$ is another set of timestamped locations and $K(\cdot)$ is a pre-defined kernel. Following~\cite{popovic2020informative}, we apply the Matérn $3/2$ kernel function, and choose kernel length scales $l_1=l_2=0.1$ spatially, and $l_3=3$ temporally, according to the sensor footprint and target speed for best approximation.

To address the issue of growing regression complexity, obsolete measurements that were acquired a long time ago are discarded, since they are not adding any meaningful information to the current agent's belief.
In practice, measurements are discarded once their intrinsic uncertainty at that timestamped location exceeds a certain threshold (in our practice, $99\%$).
That is, a fixed history of measurements is used to construct the GP, to help keep the computational complexity fixed and tractable even in infinite-horizon missions($1.993\times l_3$ for our kernel).


\section{Persistent Monitoring as a DRL Problem}

\begin{figure*}[t]
    \centering
    \vspace{-0.15cm}
    \includegraphics[width=\textwidth]{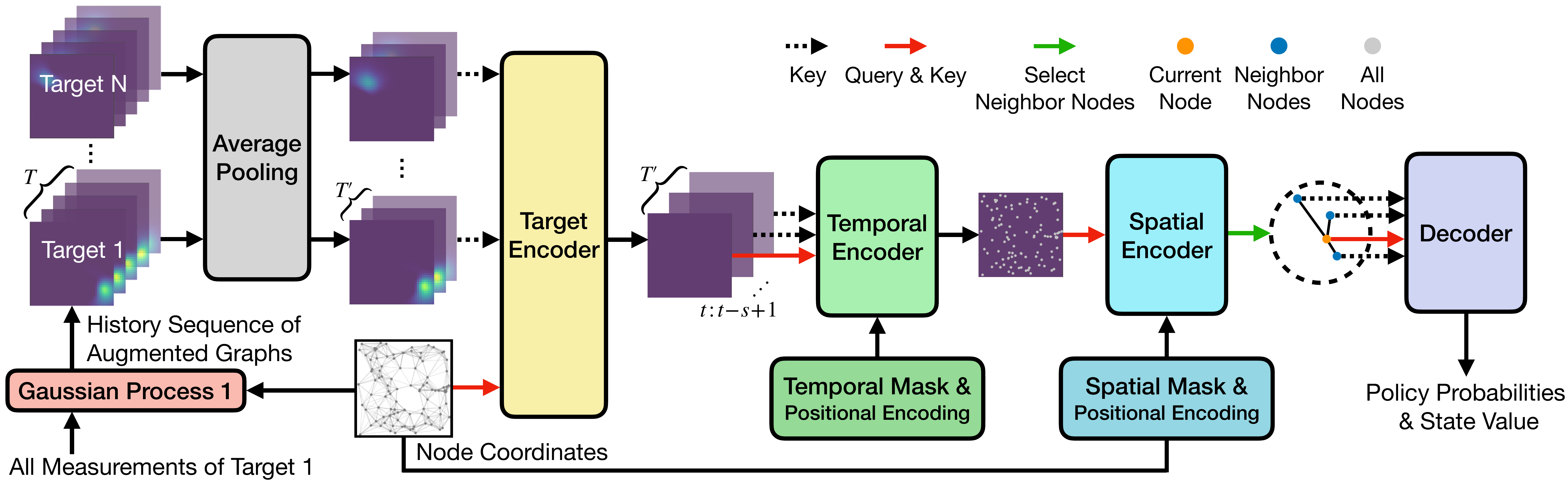}
    \vspace{-0.6cm}
    \caption{{\bf Spatio-temporal attention network.}
    The history sequence of augmented graphs, each graph being modeled/predicted by $N$ separate Gaussian Processes (one for each target), are incorporated into the target encoder before performing average pooling along the temporal axis.
    The belief of each target is integrated into the graph coordinates via cross-attention in the target encoder.
    The temporal encoder then fuses each node with all past beliefs, and the spatial encoder enhances each node by computing feature dependencies to all other nodes.
    As a result, the final policy, output by the decoder, is conditioned on the past and future beliefs of all targets as well as their spatio-temporal features.}%
    \label{fig:network}
    \vspace{-0.4cm}
\end{figure*}

This section casts persistent monitoring as a RL problem, provides a comprehensive description of our spatio-temporal attention neural network, and details our training process.

\subsection{RL Cast}

\subsubsection{\bf Graph Discretization}

In order to simplify the planning process in a continuous environment, we rely on a probabilistic roadmap~\cite{kavraki1998analysis} to discretize the space into a directed graph $G=(V,E)$, where $V \subset \mathcal{E}$ is a set of nodes/waypoints uniformly sampled from the environment.
Each node $v \in V$ is connected to its $k$ nearest neighboring nodes, forming the set of edges $E$.
This provides the agent with a discrete representation of the environment that it can use to sequentially plan its path.
That is, upon reaching the previously selected node, the agent chooses which neighboring node to visit next and moves toward it along a straight line, sequentially constructing the agent's trajectory $\psi_{0:t} = (\psi_0,\psi_1,\dots,\psi_t), \forall \psi_j \in V$.
By formulating persistent monitoring as a sequential decision-making problem, we are able to approach it with (deep) RL tools.

\subsubsection{\bf Observation}

The observation comprises the augmented graph $G'=(V',E)$ and the agent's trajectory $\psi$ within the last $T$ steps: $o_t=\{G'_{t-T+1:t}, \psi_{t-T+1:t}\}$.
Compared to the graph $G$, the augmented graph $G'$ further contains information from the agent's current belief and future prediction over all targets at each augmented node $v_j'=\left(v_j, \mu_i(v_{j,t}^*), P_i(v_{j,t}^*), \mu_i(v_{j,t+\Delta t}^*), P_i(v_{j,t+\Delta t}^*)\right) \in V', i=\{1,\dots,N\}$.
Here, $v_{j,t}^* \in V^*$ refers to the nodes coordinates $v_j$ with current timestamp $t$, and $v_{j,t+\Delta t}^*$ refers to $v_j$ with a future timestamp $t+\Delta t$ (in practice, $\Delta t=2$), following \eqref{eq:gpr} but substituting $V^*$ for $\mathcal{X^\dagger}$.
In this way, each node is enriched with information from the agent's current belief as well as future (short-term) prediction.

The final input to the agent's policy network is a sequence of the $T$ most recent augmented graphs, i.e., from time step $t-T+1$ to the current time step $t$.
In practice, we utilize average pooling with a stride and kernel size $s$ to combine temporally-adjacent augmented graph sequences as input, to leverage similarities among them.
As a result, the agent trajectory is adaptively planned, since the decision at each node is conditioned on the history of recent beliefs and future predictions of all targets.
To further facilitate learning, we use Dijkstra's algorithm~\cite{dijkstra1959note} to determine the shortest path length from each node $v_j$ to the current node $\psi_t$ through the graph, and associate it with node $v_j$.
We then associate each belief at $t'$ with trajectory length $L(\psi_{t':t})$.
This enables the agent to better infer the speed of mobile targets.

\subsubsection{\bf Action}

Each time the agent reaches a node $\psi_t$, our spatio-temporal attention network, parameterized by $\theta$, outputs a stochastic policy $\pi_\theta(v_j, (v_j,\psi_t)\in E \mid o_t)$ based on all measurements $\mathcal{Z}$ obtained recently/so far, from which we select the next neighboring node to visit (i.e., the agent's \textit{action}).
During training, we randomly sample this action according to the policy, i.e., $\psi_{t+1} \sim \pi_\theta$, but greedily select the action with the highest activation at evaluation time: $\psi_{t+1} = \arg\max\pi_\theta$.

\subsubsection{\bf Reward and Target Area}

The objective of the persistent monitoring problem is to minimize the estimation error of all mobile targets by frequently visiting them.
In~\cite{rezazadeh2020submodular}, a non-negative concave and increasing function $f(\cdot)$ is assigned to each node.
The agent collects a reward $f(\Delta t)$ each time it visits a node, where $\Delta t$ is the time interval since the last visit.
The concave nature of the reward function encourages the agent to minimize the visit interval of all targets.
Similar to their pre-defined function $f(\cdot)$, we notice that the standard-deviation of our Matérn $3/2$ kernel is also concave with respect to the distance, and therefore propose to design our rewards based on this important feature.

We first term the circular vicinity around each target its \textit{target area} $\mathcal{X}_{i,t}^\dagger$, with size identical to the sensor footprint, i.e., $\mathcal{X}_{i,t}^\dagger = \mathcal{X}^\dagger \cap S(y_{i,t})$.
In doing so, $\mathcal{X}^\dagger$ is a set of $n'$ uniformly distributed timestamped locations used to evaluate the agent's belief and uncertainty (covariance matrix trace) about a target over the domain (e.g., to estimate the agent's performance).
Thus, the agent's objective can be more precisely described as minimizing the posterior uncertainty over all (ground-truth) target areas.
We then denote the standard-deviation vector of the GP as $U$, for each $U_j = \sqrt{P_{j,j}}$, $P$ is the covariance matrix.
Hence, the average uncertainty within each target area, defined as $\overline{\sigma}_i = \sum{U_i^\dagger}/ |\mathcal{X}_{i,t}^\dagger| \in [0,1]$, is evaluated throughout the monitoring mission, where $U_i^\dagger$ is the standard-deviation inferred at target area $i$ and $|\cdot|$ the cardinality of a set.
This average uncertainty $\overline{\sigma}_i$ signifies the agent's recent presence around the target.
When the agent reaches a node (i.e., at the completion of its most-recently selected action), it will receive a positive reward signal based on the uncertainty change of all targets: $r_t = \sum_{i=1}^N{\max\left( \overline{\sigma}_{i,t-1} - \overline{\sigma}_{i,t}, 0\right)}$.
This reward incentivizes the agent to visit all mobile targets quickly and frequently.


\subsection{Network Structure}
Our proposed spatio-temporal attention neural network consists of multiple blocks of Transformer~\cite{vaswani2017attention}, sequenced into a target encoder, a temporal encoder, a spatial encoder, and finally a decoder, to capture dependencies across targets, time, and space, as shown in Fig.~\ref{fig:network}.
Our attention network is independent from the input sequence length, thus allowing our method to generalize to arbitrary numbers of targets as well as arbitrary graphs.
Each Transformer layer maps a query vector $h^q$ and a set of key-value pairs vectors $h^{k,v}$ to an output vector $h'$ by calculating a weighted sum of values.
The attention scores are derived by capturing the similarities between query and keys in each attention head:
\vspace{-0.25cm}
\begin{equation}
 \alpha_{ij} = {\rm Softmax}\left( \frac{q_i^\top\cdot k_j}{\sqrt{d}} \right), \ h'_i=\sum_{j=1}^{n}\alpha_{ij}v_{j},
\label{eq:attention}
\vspace{-0.25cm}
\end{equation}
where $q, k, v$ are derived by three separate learnable matrices in $\mathbb{R}^{d\times d}$ as $q_{i}=W^Qh^{q}_{i}, \ k_{i}=W^Kh^{k,v}_{i}, \ v_{i}=W^Vh^{k,v}_{i}$.
A residual connection and layer normalization are then applied to the output following~\cite{vaswani2017attention}.

\subsubsection{\bf Target Encoder}

Our neural network takes as input the sequence of augmented graphs $G'_{t-T+1:t}$ and first reduces its number by average-pooling (kernel size $s$, stride $s$, in practice $s=5$), leaving $T'=\ceil{T/s}$ of them.
Then, the node coordinates $V$ and the pooled augmented graph sequence are embedded respectively through two separated fully-connected layers, to output the \textit{coordinate features} $h^c$ and the \textit{target features} $h^{g_1},\dots,h^{g_N}$, for a $T'$ sequence (with time and node index omitted).
Then, these features are passed into the target encoder querying on the coordinate features (i.e., $h^q=h^c$).
By allowing the agent to output an attention score over each target, the output {\it target-aware} node features $h^{G}$ includes information about both node coordinates and GP predictions of all targets.

\subsubsection{\bf Temporal Encoder}

We denote the a sequence of node feature histories with length $T'$ as $h^{G}_1, \dots h^{G}_{T'}$ (with node index omitted), where $t'=1$ refers to the most recent pooled history $t-s+1:t$ and $t'=T'$ the last $t-T+1:t-T+s$.
As mentioned before, each node feature history at $t'$ is associated with the trajectory length $L(\psi_{t':t})$.
We use a fully-connected layer to map $L(\psi_{t':t})/c_0$ (here, $c_0=1.993\times l_3$ is a constant) to a $d$-dimensional feature $h^L_{t'}$, and add (element-wise) the corresponding node feature to yield $\tilde{h}^{G}_{t'} = h^{G}_{t'} + h^{L}_{t'}$.
This operation acts as a temporal positional encoding, which helps the agent reason about the time sequence by injecting trajectory length information.
With the most recent node features $\tilde{h}^{G}_1$ as query (i.e., $h^q=\tilde{h}^{G}_1$), $T'$ past node features are fused to produce the {\it time-aware} recent node feature $h^{TG}$ by computing a set of weighted sums in the temporal encoder.
Note that, during batch training, we apply a binary temporal mask to filter out the zero-padded inputs before an episode begins.

\subsubsection{\bf Spatial Encoder}

To relate each node's feature $h^{TG}_j$ on the graph with all other nodes, we use a spatial encoder to calculate self-attention among all nodes~\cite{cao2022catnipp}.
We encode the topology of the graph by embedding graph Laplacian eigenvectors $\lambda$ as spatial positional encoding~\cite{dwivedi2020generalization}, which enables the network to handle arbitrary graphs.
Similar to the temporal positional encoding, for each node $v_j$, we map $\lambda_j$ to $h^{E}_j$ with a fully-connected layer, and add it with the corresponding node feature: $\tilde{h}^{TG}_j = h^{TG}_j+h^{E}_j$.
Then, the the node features are enhanced in the spatial encoder by self-attention (i.e., $h^q=h^{k,v}=\tilde{h}^{TG}_j$), to finally output the {\it spatio-temporal} node features $h^{STG}$.

So far, we have obtained a $|V|\times d$ matrix that describes the feature of each node in the graph $G(V,E)$.
In doing so, each encoder, conditioned on the previous one's output, allows each spatio-temporal node feature $h^{STG}_j$ to be informed by the dependencies across all targets and their predictions, history beliefs, and features from all other nodes.

\subsubsection{\bf Decoder}

The decoder is used to yield the final policy based on the spatio-temporal node features.
We first associate each node $v_j$ with the shortest distance to the current node $\psi_t$, which is pre-solved by Dijkstra~\cite{dijkstra1959note}.
Then, a fully-connected layer maps the concatenated feature ${\rm Concat}(h^{STG}_j, {\rm dist}(v_j, \psi_t))$ to $\tilde{h}^{STG}_j$.
From these node features, we extract the current node feature (features of the current agent's node) as query (i.e., $h^q=\tilde{h}^{STG}_t$) and its neighboring nodes' features as key-value pairs $h^{k,v}$ for the decoder unit.
There, inspired by the Pointer Network~\cite{vinyals2015pointer}, we directly use the decoder's attention score $\alpha_j$ as the final output policy $\pi_\theta$, from which the agent chooses the next node to move to.
This scheme relaxes the requirement of a fixed policy size, instead adapting the policy's dimension dynamically to the number of neighboring nodes.
Together with spatial positional encoding, this endows our network with the ability to generalize to arbitrary graphs and topologies.


\subsection{Network Training}
We use the same network to output a state value estimate $\hat{V^s}(s_t;\theta)$, and calculate the advantage as $\hat{A}_t=R(t;\theta)-\hat{V^s}(s_t;\theta)$, where $R(t;\theta)=\sum_{t'=0}^t{\gamma^{t'}r(t';\theta)}$ is the discounted return ($\gamma=0.99$ in practice).
We adopt the PPO algorithm to train our neural network~\cite{schulman2017proximal}, with following loss minimized to optimize the policy:
\vspace{-0.25cm}
\begin{equation}
    L(\theta) = \mathbb{E}_t\left[\min _{\theta}(\rho_t(\theta)\hat{A}_t, \operatorname{clip}\left(\rho_t(\theta) , 1-\epsilon, 1+\epsilon\right)\hat{A}_t)\right],
\vspace{-0.25cm}
\end{equation}
where $\operatorname{clip}(\cdot)$ and the ratio $\rho$ are defined in~\cite{schulman2017proximal}.

During training, we assume the exact number of targets and their initial positions are known to the agent as prior information as $x_{i,0}^*=\{y_{i,0},0\}$ with $z_{i,0}=1$).
The targets move in a 2D unit square $[0,1]^2$ in a set of pre-solved TSP patterns, each consisting of 50 randomly generated nodes (unknown to the agent).
For each training batch, the number of nodes in the graph is uniformly randomized within $[100,200]$ with 10 nearest neighboring nodes connected.
The range of history sequence length is randomly drawn within $[50,100]$, and the number of targets in $[2,5]$.
While persistent monitoring is an infinite-horizoned problem, we terminate each episode once it exceeds 256 decision steps to allow for episodic training.
The speed ratio $r_v$ is randomized in the range $[0,r'_v]$, where $r'_v$ gradually increases from 0 to 0.1 over the first 10,000 episodes (curriculum learning approach).
A set of measurement $\mathcal{Z}_t$ is obtained once the agent traveled 0.1 from the last measurement.
We train our neural network with a minibatch size of 512, and 4 runs for each batch.
The learning rate starts from $10^{-4}$ and decays every 64 episodes by a factor of 0.96.
We trained our model on a server with 4 NVIDIA RTX 3090 GPUs and a i9-10980XE CPU (note that similar performance can be reached using a single GPU with a smaller minibatch size).
We run 32 instances in parallel to accelerate data collection and train our model for a total of 50,000 episodes, which takes approximately 30 hours\footnote{Our full code will be released upon paper acceptance.}.

\begin{table*}[t]
    \centering
    \caption{{\bf Comparison with lawnmower coverage and TSP loop (100 instances each).}
    We report the overall average uncertainty at each target area (standard deviation of uncertainty between targets in parentheses), minimum times of observation among all targets, and the average JS divergence (estimate error) to the true distribution.}
    \vspace{-0.15cm}
    \begin{tabular}{c|c@{\hspace{1.5\tabcolsep}}c@{\hspace{1.5\tabcolsep}}c@{\hspace{1.5\tabcolsep}}c|cccc|cccc}
    \toprule
        \multicolumn{1}{c|}{Metric} & \multicolumn{4}{c|}{\bf Unc} & \multicolumn{4}{c|}{\bf MinOb} & \multicolumn{4}{c}{\bf JSD} \\
        \hline
        \multicolumn{1}{c|}{Speed Ratio $r_v$} & 1/30 & 1/20 & 1/10 & 1/7 & 1/30 & 1/20 & 1/10 & 1/7 & 1/30 & 1/20 & 1/10 & 1/7 \\
        \midrule
        \multicolumn{1}{c|}{Method} & \multicolumn{12}{c}{Target number $N=2$} \\
        \hline
        Lawnmower & 0.838(0.078) & 0.841(0.077) & 0.841(0.085) & 0.846(0.082) & 7.06 & 7.50 & 7.45 & 6.71 & 0.223 & 0.259 & 0.346 & 0.392 \\
        TSP Loop & 0.643(0.029) & {\bf 0.640}(0.038) & 0.800(0.075) & 0.872(0.067) & {\bf 59.24} & {\bf 60.04} & 28.86 & 16.00 & {\bf 0.070} & {\bf 0.094} & 0.288 & 0.365 \\
        Ours  & {\bf 0.634}(0.063) & 0.644(0.073) & {\bf 0.718}(0.107) & {\bf 0.766}(0.116) & 46.55 & 46.86 & {\bf 32.94} & {\bf 22.41} & 0.109 & 0.131 & {\bf 0.242} & {\bf 0.312} \\
        \midrule
        \multicolumn{1}{c|}{Method} & \multicolumn{12}{c}{Target number $N=4$} \\
        \hline
        Lawnmower & 0.839(0.119) & 0.841(0.121) & 0.844(0.123) & 0.848(0.124) & 6.17 & 6.21 & 5.68 & 5.51 & 0.223 & 0.257 & 0.347 & 0.392 \\
        TSP Loop & {\bf 0.669}(0.063) & {\bf 0.724}(0.096) & 0.869(0.113) & 0.894(0.103) & {\bf 20.80} & 15.36 & 4.62 & 4.07 & {\bf 0.083} & 0.165 & 0.345 & 0.391 \\
        Ours & 0.707(0.102) & 0.725(0.115) & {\bf 0.779}(0.133) & {\bf 0.807}(0.131) & 17.72 & {\bf 15.67} & {\bf 11.19} & {\bf 9.08} & 0.125 & {\bf 0.164} & {\bf 0.290} & {\bf 0.357} \\
        \midrule
        \multicolumn{1}{c|}{Method} & \multicolumn{12}{c}{Target number $N=6$} \\
        \hline
        Lawnmower & 0.837(0.131) & 0.840(0.133) & 0.842(0.135) & 0.844(0.136) & 5.53 & 5.73 & 5.35 & 4.85 & 0.223 & 0.259 & 0.345 & 0.393 \\
        TSP Loop & {\bf 0.701}(0.084) & 0.775(0.119) & 0.864(0.123) & 0.882(0.118) & {\bf 15.80} & 6.84 & 3.54 & 3.18 & {\bf 0.096} & 0.209 & 0.348 & 0.390 \\
        Ours & 0.743(0.118) & {\bf 0.759}(0.128) & {\bf 0.804}(0.132) & {\bf 0.822}(0.131) & 9.51 & {\bf 8.47} & {\bf 6.32} & {\bf 5.73} & 0.146 & {\bf 0.192} & {\bf 0.311} & {\bf 0.366} \\
        \midrule
        \multicolumn{1}{c|}{Method} & \multicolumn{12}{c}{Target number $N=8$} \\
        \hline
        Lawnmower & 0.839(0.135) & 0.840(0.137) & 0.842(0.141) & 0.843(0.142) & 5.27 & 5.21 & 4.94 & {\bf 4.88} & 0.224 & 0.258 & 0.346 & 0.390 \\
        TSP Loop & {\bf 0.731}(0.101) & 0.803(0.125) & 0.856(0.128) & 0.871(0.126) & {\bf 10.37} & 3.92 & 3.40 & 3.28 & {\bf 0.118} & 0.235 & 0.344 & 0.388 \\
        Ours & 0.761(0.125) & {\bf 0.776}(0.131) & {\bf 0.811}(0.134) & {\bf 0.828}(0.132) & 5.99 & {\bf 5.92} & {\bf 4.95} & 4.74 & 0.157 & {\bf 0.202} & {\bf 0.315} & {\bf 0.372} \\
    \bottomrule
    \end{tabular}
    \label{tab:comparison}
    \vspace{-0.4cm}
\end{table*}

\begin{figure}[t]
    \centering
    \subcaptionbox{$N=4,r_v=1/20$}{\vspace{-0.15cm}
    \includegraphics[width=0.22\textwidth]{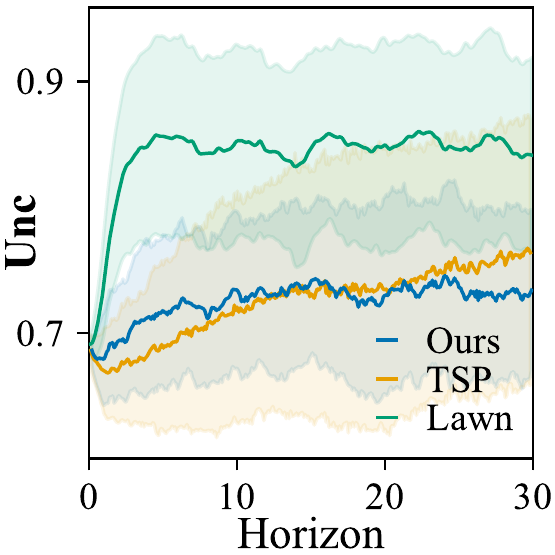}}%
    \hfill
    \subcaptionbox{$N=6,r_v=1/20$}{\vspace{-0.15cm}
    \includegraphics[width=0.22\textwidth]{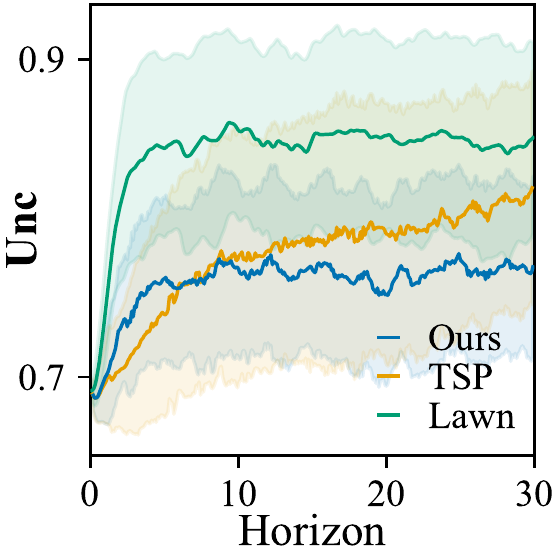}}%
    \hfill
    \subcaptionbox{$N=4,r_v=1/20$}{\vspace{-0.15cm}
    \includegraphics[width=0.48\textwidth]{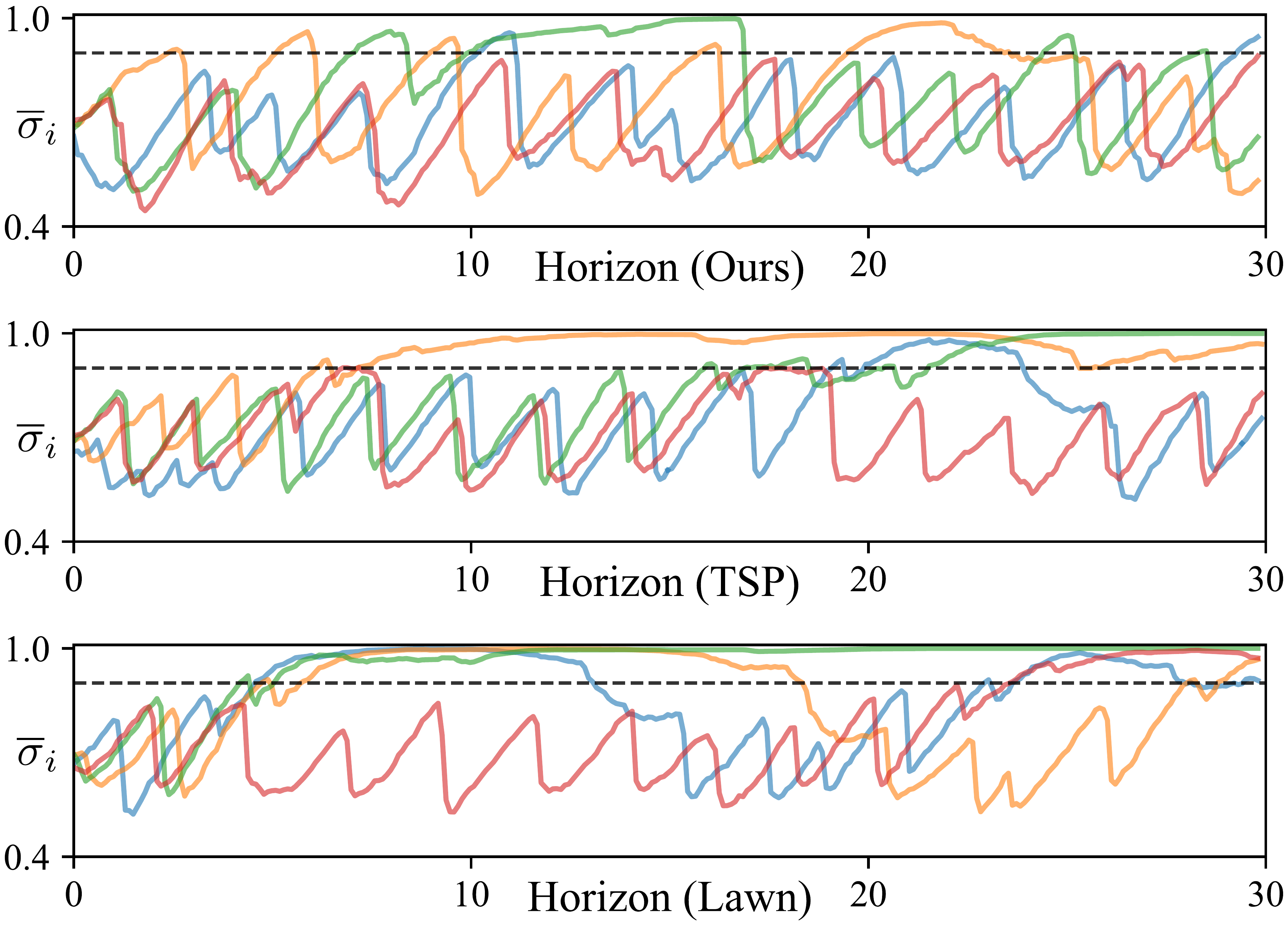}}%
    \vspace{-0.15cm}
    \caption{{\bf Comparison with Lawnmower coverage and TSP loop (100 instances each).}
    (a, b) Average uncertainty throughout the mission (4 and 6 targets).
    (c) Uncertainty over each target area over one example episode (4 targets).}
    \vspace{-0.45cm}
    \label{fig:traj}
\end{figure}


\section{Experiments}

In this section, we conduct a series of experiments to compare our method with a TSP-based baseline (termed TSP loop) as well as standard Boustrophedon/lawnmower coverage.
We investigate the performance of our method with various graph/history sizes and report their performance through an ablation study.
Additionally, we show the generalization performance of our method in scenarios without any initial prior, and finally validate our model in a drone-based AirSim simulation to monitor a set of mobile targets.


\subsection{Comparisons Analysis}
We report the comparison results of our method with Boustrophedon coverage~\cite{choset2000coverage} and TSP loop derived from~\cite{yu2017optimal} in a number of scenarios.
The Boustrophedon method simply performs uniform coverage following a {\it lawnmower} pattern (i.e., move back and forth) over the monitoring domain.
It plans a pre-defined path which is not related to the measurements and do not focus on any specific regions (i.e., \textit{non-adaptive} coverage).
On the other hand, our TSP-based baseline is adaptive, and plans a {\it TSP loop} based on all last-seen target positions, sequentially moving each target in a straight line from its last-seen position (along its last-seen heading).
Whenever a target is observed, the target's last-seen position and subsequently, the TSP loop is updated.
We compare our method (with $|V|=200, T=100$) against these two baselines in a fixed set of randomly generated environments while keeping all configurations identical.
There, the mission horizon is fixed to a total trajectory length $L(\psi)$ of 30.
As it is an essential requirement for the TSP loop baseline, the exact number of targets and their initial locations are given as a priori to the agent in all baselines.

We evaluate the monitoring performance using the following three metrics:
(1)~{\bf Unc} $\sum_{i}^{N}{\overline{\sigma}_i}/N$, which is the overall average uncertainty in all target areas (lower is better).
We also report the standard deviation between each target's average uncertainty.
(2)~{\bf MinOb} $\min_i{\sum_{t}{z_{i,t}}}$, which is the minimum number of target observations obtained among all targets (higher is better).
(3)~{\bf JSD} $\sum_{i}^{N}{\operatorname{JSD}(\mu_i(\mathcal{X}^\dagger) \parallel Y_i(\mathcal{X}^\dagger))}/N$, which is the average Jensen–Shannon divergence between the agent's belief over each target and that target's interpolated ground truth (i.e., where the true target position is being turned into a 2D Gaussian distribution $Y_i$ by placing a normalized Gaussian (peak value 1) centered at $y_{i,t}$ with standard deviation identical to the agent's sensor $l_1,l_2$ (i.e., 0.1)).
{\bf JSD} is used to measure the average distance and estimate error between two distributions (lower is better).

\begin{table}[t]
    \centering
    \caption{{\bf Performance of various numbers of graph nodes and history beliefs of our method and its ablation (100 instances each).} We report the {\bf JSD} of those variants.
    The ``200 (w/o)'' line refers to the ablation variant of our method $|V|=200$ without future predictions input, while the ``LSTM'' model replaces our temporal encoder with an LSTM cell.}
    \vspace{-0.15cm}
    \begin{tabular}{c|cccccc}
    \toprule
         \diagbox[width=5em, height=2em]{$T$}{$|V|$} & 50 & 100 & 150 & 200 & 300 & 400 \\
         \midrule
         1   & 0.313 & 0.314 & 0.322 & 0.322 & 0.326 & 0.329 \\
         25  & 0.301 & 0.296 & 0.297 & 0.292 & 0.304 & 0.296 \\
         50  & 0.299 & \textbf{0.291} & 0.292 & 0.297 & 0.293 & \textbf{0.287} \\
         100 & 0.301 & 0.299 & 0.295 & \textbf{0.290} & \textbf{0.290} & 0.293 \\
         200 & 0.299 & 0.298 & 0.292 & 0.295 & 0.292 & \textbf{0.291} \\
         \midrule
         200 (w/o) & 0.305 & 0.305 & 0.301 & 0.300 & 0.303 & 0.305 \\
         LSTM & 0.306 & 0.303 & 0.311 & 0.313 & 0.318 & 0.322 \\
    \bottomrule
    \end{tabular}
    \label{tab:variants}
    \vspace{-0.5cm}
\end{table}

As shown in Table~\ref{tab:comparison}, in relatively simple scenarios (i,e., lower target speeds), TSP loop can achieve near-optimal performance and easily outperforms our method.
Since the persistent monitoring of a set of low-speed targets is well-approximated to standard TSP, it is expected that TSP loop can become the optimal solution.
However, as the target speed and number of targets increase, our method outperforms all other baselines.
We notice that the performance of TSP loop drops drastically once the speed ratio goes beyond 1/20, being outperformed by our method, even though our agent is confined to a graph and the TSP loop agent is not.
Upon closer inspection, given accurate prior, TSP loop performs the best at early stages of a mission, but its {\bf Unc} gradually increases as the monitoring progresses and finally exceeds our method, see Fig.~\ref{fig:traj}(a, b).
This phenomenon suggests that TSP loop is more likely to lose targets without being able to relocate them during the mission, highlighting an important limitation for infinite-horizon monitoring tasks.
The lawnmower coverage, as a non-adaptive method, performs the worst among almost all scenarios, except when the environment is highly dynamic ($N=8,r_v=1/7$), where it naturally outperforms the other approaches in terms of addressing all targets equally ({\bf MinOb}).

In our experiments, a target is defined as lost if $\overline{\sigma}_i > 0.9$, as shown in the horizontal dashed line in Fig.~\ref{fig:traj}(c).
TSP loop solely relies on the last-seen target positions, and if a target does not appear in its sensor footprint when attempting to relocate it, the agent will directly skip it and travel to the next target's last-seen position.
Therefore, once a target is lost, TSP loop struggles at relocating it and can only ever really do it out of luck.
To boost the performance of TSP loop, we also attempted to replace the last-seen target position with the its future predicted location from the GP regression.
However, performance remained similar, except that future predictions increased the performance of experiments with $r_v=1/30$ by a small margin.
Our method, on the other hand, considers the spatio-temporal behaviors of the targets and implicitly decide which area to go to both for short- and longer-term monitoring.
We believe that this learned ability to reason about spatio-temporal data is at the source of our improved performances, especially in complex scenarios.

\begin{figure}[t]
    \centering
    \subcaptionbox{$N=4,r_v=1/20$}{\vspace{-0.15cm}
    \includegraphics[width=0.22\textwidth]{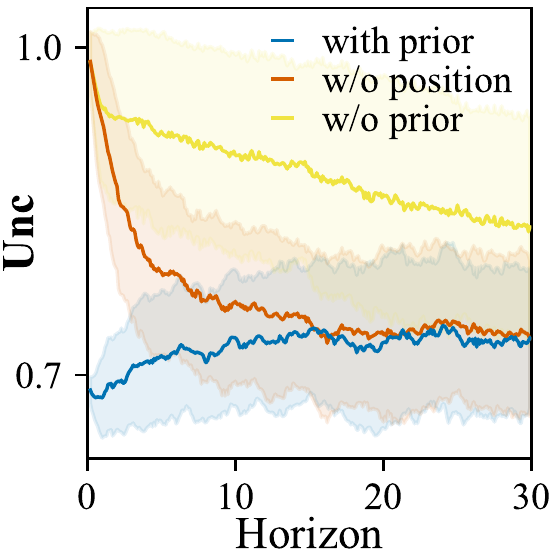}}%
    \hfill
    \subcaptionbox{$N=6,r_v=1/20$}{\vspace{-0.15cm}
    \includegraphics[width=0.22\textwidth]{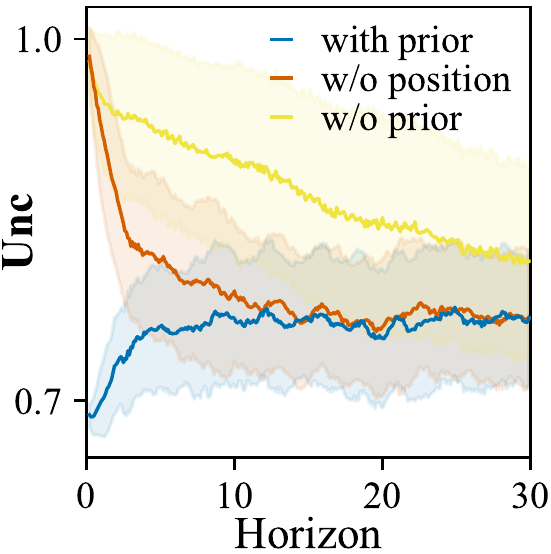}}%
    \vspace{-0.2cm}
    \caption{{\bf Comparison with between prior conditions (100 instances each).}
    The average uncertainty throughout the monitoring horizon given prior, only target number, or no prior information, with 4 and 6 targets.}
    \vspace{-0.5cm}
    \label{fig:noprior}
\end{figure}


\subsection{Variants and Ablation Analysis}
We further vary the number of nodes and history sequence and evaluate their performance using the same trained model.
Additionally, we further train two variants of our model, as an ablation study to verify the effectiveness of our different techniques.
We train one model without future predictions input (i.e., $\mu_i(v_{j,t+\Delta t}^*)$ and $P_i(v_{j,t+\Delta t}^*)$), and in the other model, we replaced the temporal encoder with a long short-term memory (LSTM) unit instead.
The LSTM unit is fed with $\tilde{h}^{STG}_j$ at each decision step, and produces the output and hidden/cell states for memory.
The results of these variants are presented in Table~\ref{tab:variants} keeping $N=4, r_v=1/10$, where we highlight 5 results with the least {\bf JSD} error.

We note that our method exhibits the poorest performance when the graph lacks adequate discretization or is devoid of past beliefs.
With an increase in the number of nodes and history sequence length, the {\bf JSD} experiences a gradual decline, which becomes marginal once provided with sufficient graph fineness and history beliefs.
Through the comparison of our method with and without future prediction, we conclude that the incorporation of future predictions is highly likely to enhance our performance, as it provides additional information on the agent's belief.
The performance of the LSTM-only agent generally lies in-between $T=1$ and $T=25$, which is possibly due to the LSTM's lesser ability at retaining a long-term information and establish meaningful connections among them.
Through these two comparison groups, we validate the effectiveness of the future prediction and our temporal encoder for modelling longer-term dependencies given enough history sequence length.

\begin{figure}[t]
  \centering
  \setlength{\belowcaptionskip}{-0.3cm}
  \includegraphics[width=0.7\linewidth]{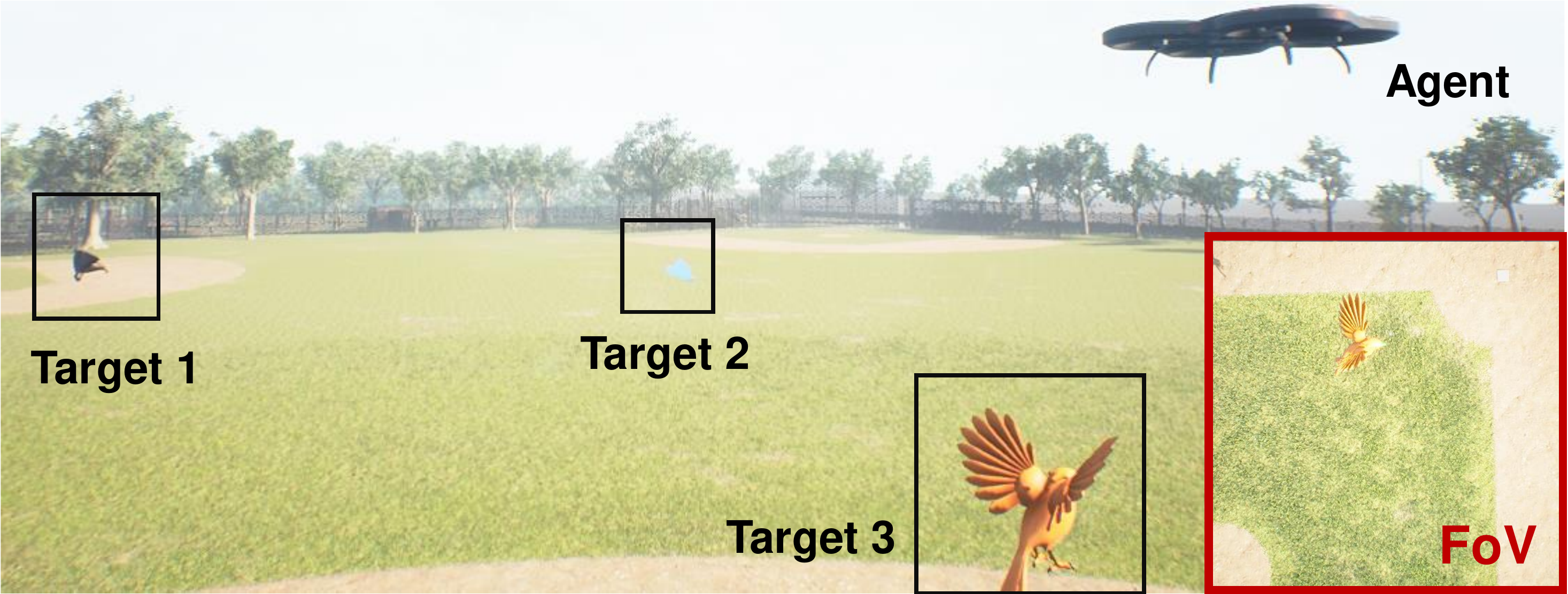}
  \vspace{-0.1cm}
  \caption{{\bf Airsim simulation experiment}.
  The drone (agent) is tasked to monitor three different animals (in black squares).
  The locations of these animals are calculated from images taken by the downward-facing camera of the drone (red square at bottom-right)}
  \label{fig:airsim}
  \vspace{-0.4cm}
\end{figure}


\subsection{Generalization to Scenarios without Prior}

TSP loop requires (accurate) prior knowledge over the number and locations of targets.
Our method, relying on attention mechanism, should theoretically accommodate varying sequence.
Hence, we investigate the efficacy of our model (trained with prior information) in two scenarios, where the agent has access to (1)~only the total number of target, but not their initial position, or (2)~no information at all.
The results are presented in Fig.~\ref{fig:noprior}.

We note that, in the first scenario, the agent can quickly cover the domain at the initial stages of the mission, and exhibit similar performance once all targets are found (after a horizon of around 15 in this case).
In the second scenario, whenever a new target is found, we allow the agent to create an additional augmented graph, to be added to the network inputs.
There, we observe that the rate at which the {\bf Unc} decreases is significantly lower compared to the first scenario, but still shows that our agent is slowly gathering information (finding targets), although it does not locate all of them by the end of the mission in this case.
This confirms that the number of target is a determinant factor of the agent's decision-making and final performance.
However, we also note that our method demonstrates a form of natural robustness even when the initial target locations (and to a lesser extent, their number) are not known a priori, highlighting its ability to recover from target loss and to generalize to complex scenarios without any prior information.


\subsection{High-Fidelity Simulation and Attention Visualization}

Fig.~\ref{fig:airsim} shows a static frame from our high-fidelity simulation environment using AirSim simulator, where we rely on our trained model to let a drone monitor a set of random ground targets (here, walking animals) using a downward-facing camera.
During the simulation, a photo is captured by the downward-facing camera at certain intervals as the drone moves along its path.
The captured photo is subsequently processed to detect the possible presence of each target, and the resulting agent belief is used by our trained model, which only takes approximately 0.1 seconds for each decision (forward inference).
Subsequently, our model assigns the next waypoint for the drone, in order to sequentially visit and continually re-locate all targets.

Finally, to better understand the functionality of each encoder in our network architecture, we also present a visualization of the attention score of each head in relevant scenarios.
These two additional visual results are included in our supplementary video.


\section{Conclusion}

In this work, we introduce a deep reinforcement learning approach to address the persistent monitoring problem.
To monitor a group of targets moving in a given domain, our proposed method leverages attention mechanisms to learn the spatio-temporal dependencies across targets, thus allowing agent to plan a context-aware trajectory implicitly informed by the past measurements and short-term future predictions over the location of all known targets.
Our experiments demonstrate the superiority of our proposed spatio-temporal attention network over two conventional baselines, particularly in complex scenarios where targets move rapidly compared to the agent (but not too rapidly, where the optimal strategy seems to perform uniform coverage).
Moreover, we highlight the natural generalization/adaptability of our model to scenarios where prior information is incomplete/inaccurate, suggesting that our approach can potentially extend its applicability to a wider range of real-world monitoring tasks without any additional retraining, where the number and locations of targets may be unknown a priori or subject to frequent changes.

Future work will look at multi-agent, multi-target persistent monitoring, where agents need to collectively minimize estimation error through cooperation.
We also plan to develop/implement our model on-robot, for deployments in cases where the environment may be partially observable due to the presence of obstacles blocking line-of-sight.


\section*{ACKNOWLEDGMENT}

This work was supported by Temasek Laboratories (TL@NUS) under grants TL/SRP/21/19 and TL/FS/2022/01.

\bibliography{ref}

\end{document}